\documentclass[conference,a4paper]{IEEEtran}
\IEEEoverridecommandlockouts

\usepackage[hidelinks]{hyperref}
\usepackage[cmex10]{amsmath}%American Math Society(AMS) math formatting
\usepackage{amssymb,amsfonts}%AMS extra symbols and fonts
\usepackage{dblfloatfix}%fix double column figure ordering and placement
\usepackage{multirow}
\usepackage{booktabs} % 用于 \toprule, \midrule 和 \bottomrule

\usepackage[ruled,vlined]{algorithm2e}
\usepackage{caption}
\usepackage{float}  % 在导言区添加此行
\usepackage{graphicx}

\graphicspath{{Figures/PDF/}{Figures/PNG/}}

\usepackage{booktabs}
\usepackage{siunitx}
\usepackage[numbers,compress]{natbib}
\usepackage{texnames}
\usepackage{bm,bbm}
\usepackage{orcidlink}

\begin{document}

\title{\uppercase{PSRGS:Progressive Spectral Residual of 3D Gaussian for High-Frequency Recovery}
}

\author{
    \IEEEauthorblockN{BoCheng Li\orcidlink{0009-0000-7543-2914}}
	\IEEEauthorblockA{\textit{Aerospace Information} \\ \textit{ Research Institute}\\
		101408 BeiJing, China\\
		libocheng241@mails.ucas.ac.cn}
	\and
	\IEEEauthorblockN{WenJuan Zhang\orcidlink{0000-0002-0534-0974}}
	\IEEEauthorblockA{\textit{Aerospace Information } \\ \textit{ Research Institute}\\
        101408 BeiJing, China\\
		zhangwj@radi.ac.cn}
	\and
	\IEEEauthorblockN{Bing Zhang\orcidlink{0000-0002-9185-1382}}
	\IEEEauthorblockA{\textit{Aerospace Information Research Institute} \\
    \textit{University of Chinese Academy of Sciences}\\
        101408 BeiJing, China\\
		zhangbing@aircas.ac.cn}
    \and
    %\hspace{1cm} % 手动控制起始位置
	\IEEEauthorblockN{YiLing Yao\orcidlink{0009-0009-1025-8674}}
    \hspace{10cm} % 手动控制起始位置
	\IEEEauthorblockA{\textit{Aerospace Information}  \textit{ Research Institute}\\
        101408 BeiJing, China\\
		yaoyiling22@mails.ucas.ac.cn}
    \and
    \IEEEauthorblockN{YaNing Wang\orcidlink{0009-0000-1801-904X}}
    \hspace{1cm} % 手动控制起始位置
	\IEEEauthorblockA{\textit{Aerospace Information} \textit{ Research Institute}\\
        101408 BeiJing, China\\
		yaningstu@163.com}
}

\maketitle
\begin{abstract}
	3D Gaussian Splatting (3D GS) achieves impressive results in novel view synthesis for small, single-object scenes through Gaussian ellipsoid initialization and adaptive density control. However, when applied to large-scale remote sensing scenes, 3D GS faces challenges: the point clouds generated by Structure-from-Motion (SfM) are often sparse, and the inherent smoothing behavior of 3D GS leads to over-reconstruction in high-frequency regions, where have detailed textures and color variations. This results in the generation of large, opaque Gaussian ellipsoids that cause gradient artifacts. Moreover, the simultaneous optimization of both geometry and texture may lead to densification of Gaussian ellipsoids at incorrect geometric locations, resulting in artifacts in other views. To address these issues, we propose PSRGS, a progressive optimization scheme based on spectral residual maps. Specifically, we create a spectral residual significance map to separate low-frequency and high-frequency regions. In the low-frequency region, we apply depth-aware and depth-smooth losses to initialize the scene geometry with low  threshold. For the high-frequency region, we use gradient features with higher threshold to split and clone ellipsoids, refining the scene. The sampling rate is determined by feature responses and gradient loss. Finally, we introduce a pre-trained network that jointly computes perceptual loss from multiple views, ensuring accurate restoration of high-frequency details in both Gaussian ellipsoids geometry and color. We conduct experiments on multiple datasets to assess the effectiveness of our method, which demonstrates competitive rendering quality, especially in recovering texture details in high-frequency regions. 
\end{abstract}


\section{Introduction}
\label{sec:introduction}

\begin{figure}[!ht]
	\centering
	\includegraphics[width=\linewidth,keepaspectratio]{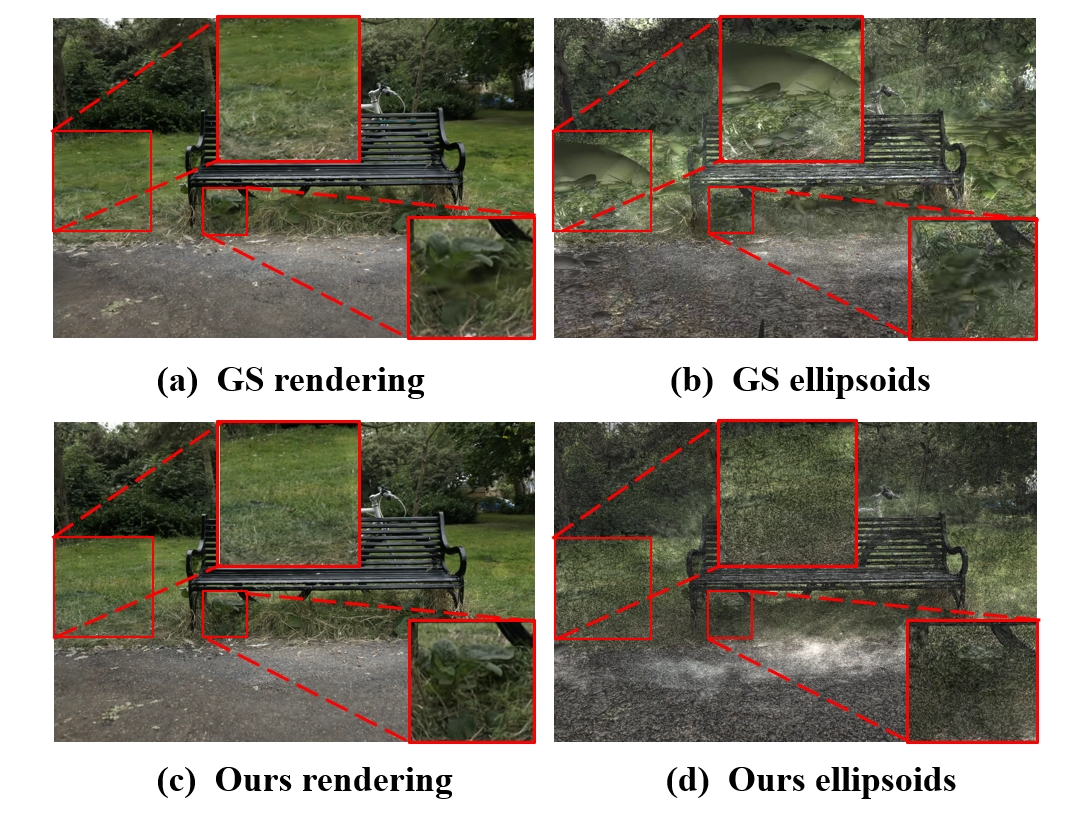}
	\caption{The over-reconstruction tendency of 3D GS tends to generate large Gaussian ellipsoids, leading to artifacts in the scene. In contrast, our method manages the geometric structure effectively, reconstructing the scene more accurately and eliminating artifacts}
	\label{fig:arc}
  \end{figure}

\begin{figure}[!ht]
    \centering
    \includegraphics[width=\linewidth, keepaspectratio]{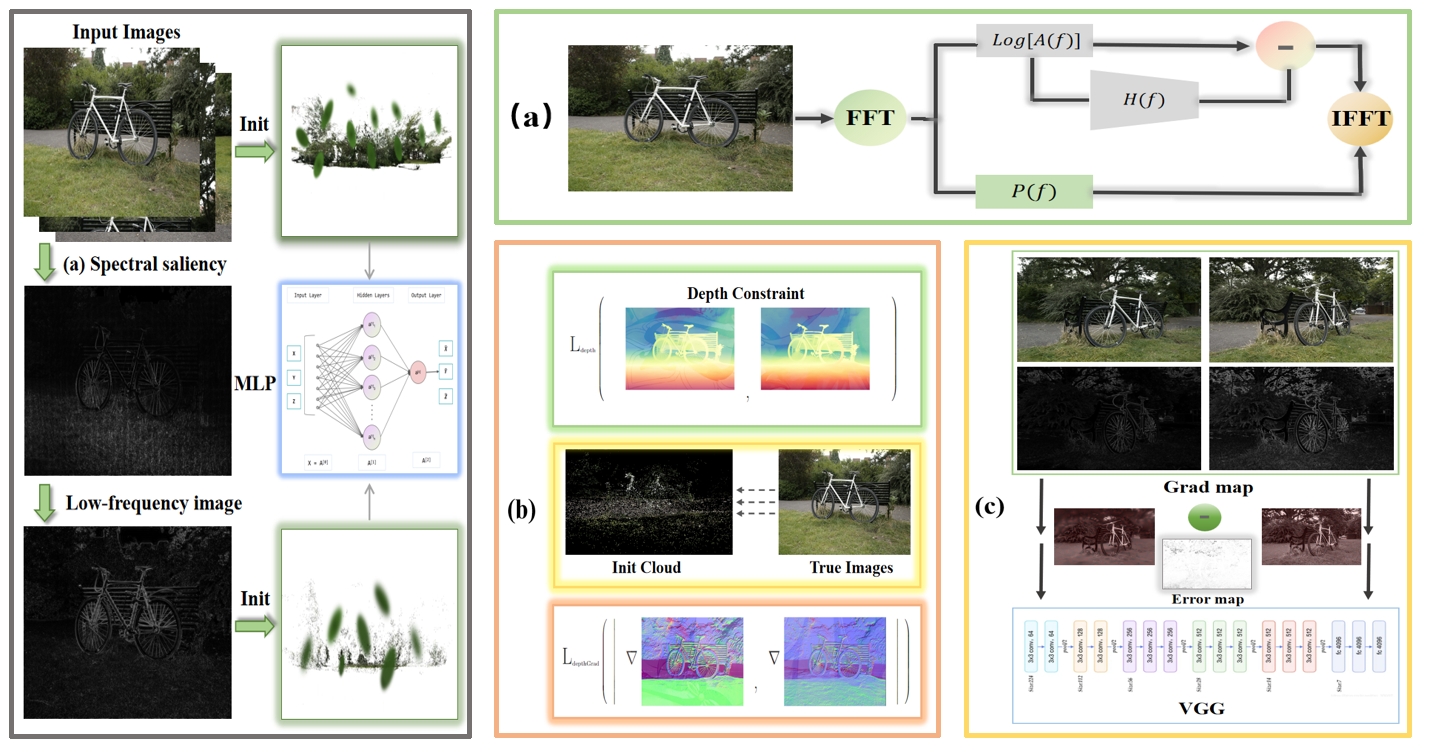}
    \caption{\textbf{Overview of Our PRSGS Pipeline.} (a) \textbf{Spectral Salient Residual Decoupling.} We perform a spectral transform, where the smoothed residual of the logarithmic magnitude spectrum is combined with the original phase spectrum to generate the salient map. The standard deviation for filtering is learned by an MLP network. (b) \textbf{Low-Frequency Processing Stage.} A higher threshold is applied for densification, and depth perception loss and smoothness loss are used to constrain the scene's depth and normal maps. (c) \textbf{High-Frequency Processing Stage.} Gradient feature maps of the image are generated, and densification is performed with a lower threshold along the gradient direction. Additionally, a pre-trained network is used to compute the multi-view geometric perception loss, incorporating plane constraints to reduce the generation of erroneous Gaussian points.}
    \label{fig:workflow}
\end{figure}

In the fields of computer vision and computer graphics, novel view synthesis has always been an important and challenging task. The synthesis of new views from images captured at known viewpoints has a wide range of potential applications, such as in virtual reality, augmented reality, and autonomous driving. In recent years, deep learning-based methods, such as Neural Radiance Fields (NeRF)\citep{mildenhallNeRFRepresentingScenes2020}, have made significant progress in novel view synthesis. However, NeRF requires sampling through a large number of rays, which results in rendering speeds that do not meet real-time requirements.

Unlike NeRF, 3D Gaussian Splatting (3D GS)\citep{kerbl3DGaussianSplatting2023a} uses 3D point clouds obtained from Structure-from-Motion (SfM) and parameterizes their shape, position, and texture properties with Gaussian ellipsoids. The application of adaptive density control ensures accurate scene representation, leading to promising results in novel view synthesis. However, 3D GS has limitations when dealing with high-frequency regions, especially in large-scale remote sensing scene images, the Earth's surface includes a variety of land cover types such as vegetation, water, soil, roads, and buildings, resulting in remote sensing images with rich land cover details and significant variations across different regions. However, 3D GS have relatively weak reconstruction capabilities in high-frequency areas, making it difficult to directly apply to remote sensing images \citep{zhiwenyanMultiScale3DGaussian2023,zhangFreGS3DGaussian2024,xiongSparseGSRealtime360deg2024}. Specifically, high-frequency regions typically contain more information, manifesting as rich textures and colors, the radiometric and geometric features of adjacent pixels  change drastically, with rapid changes in pixel values and gradients, making them more complex than low-frequency regions \citep{mijeongkim4DGaussianSplatting2024,haolinli3DHGS3DHalfGaussian2024}. As a result, the corresponding 3D point clouds should be denser to ensure that the trained Gaussian ellipsoids can accurately simulate the rendered view. In contrast, point clouds generated by SfM are often sparse\citep{ruihongyinFewViewGSGaussianSplatting2024,gaochaosongGVKFGaussianVoxel2024}, and 3D GS applies the same thresholding to both high-frequency and low-frequency regions\citep{yeAbsGSRecoveringFine2024}. With large scenes, the increased scene scale reduces the spatial proportion of high-frequency regions\citep{jiaqilinVastGaussianVast3D2024}, and the Gaussian ellipsoids in small-scale high-frequency regions are insufficient to recover the complete details. Furthermore, optimization of the error function may result in some incorrectly positioned ellipsoids causing opaque artifacts, especially in complex scenes. 3D GS also inherits the cumulative errors of SfM point clouds, and due to the distribution of loss gradients, incorrect ellipsoids may continue to enlarge\citep{yangLocalizedGaussianPoint2024,yangfuCOLMAPFree3DGaussian2024}.

The emergence of this issue can be attributed to the simultaneous optimization of both geometry and texture in the prior training process of 3D GS, where part of the texture loss is absorbed by the geometry loss, leading to geometric errors in the Gaussian ellipsoids. Previous works\citep{yeAbsGSRecoveringFine2024,chengGaussianPro3DGaussian2024, yaoRSGaussian3DGaussian2024} have proposed various methods to address this problem. Ye et al. \citep{yeAbsGSRecoveringFine2024} incorporates absolute value threshold segmentation gradient loss, but it cannot constrain the correctness of their geometric positioning. Cheng et al. \citep{chengGaussianPro3DGaussian2024} introduces a multi-view patch matching strategy. This approach effectively fixes the geometry but has limited impact on restoring texture details in high-frequency regions. Yao et al. \citep{yaoRSGaussian3DGaussian2024} introduces additional ground truth point clouds, guiding densification along the correct direction, but it requires significant prior knowledge.

In this paper, we propose a progressive scene construction framework that decouples 3D scenes into low-frequency structural geometry and high-frequency texture details using spectral residual maps, optimizing each part separately to achieve more accurate scene representation. Our method does not require additional priors, and the decoupled optimization prevents texture loss from propagating into the geometry loss. In this way, our approach effectively restores high-frequency details, achieving superior results.

\section{METHOD}
\label{sec:method}

We propose PSRGS, a progressive approach for handling scene geometry and texture. Figure \ref{fig:workflow} illustrates the workflow of PSRGS. In Section \ref{sec:preliminary}, we will introduce the related work on 3D GS, followed by a detailed description of the three main modules in our work: the low-frequency and high-frequency region decoupling module, the low-frequency geometry constraint module, and the high-frequency texture recovery module.

\begin{table*}[!ht]
    \centering
	\renewcommand{\arraystretch}{0.8} % 缩小行高
    \caption{Comparison of metrics across different methods. Higher PSNR and SSIM are better, while lower LPIPS is better.}
    \label{table:comparison1}
    \footnotesize % 设置字体大小为 footnotesize
    \resizebox{0.9\textwidth}{!}{%
    \begin{tabular}{ccccccccc}
    \toprule
    \multicolumn{2}{c}{\textbf{Dataset}} & \multicolumn{5}{c}{\textbf{MipNeRF360}} & \multicolumn{2}{c}{\textbf{LUND}} \\  
    \cmidrule(lr){1-2} \cmidrule(lr){3-7} \cmidrule(lr){8-9}  
    Metrics & Methods & Bicycle & Bonsai & Garden & Flowers & Stump & Cap & Blood \\ 
    \midrule
    
    \multirow{5}{*}{SSIM $\uparrow$} 
        & 3D GS \cite{kerbl3DGaussianSplatting2023a}  & 0.732  & 0.923  & 0.840  & 0.614  & 0.720  & 0.722  & 0.912  \\ 
        & ABSGS \cite{yeAbsGSRecoveringFine2024} & 0.752 & 0.944 & 0.860 & 0.609 & 0.761 & 0.788 & 0.918 \\
        & RSGS \cite{yaoRSGaussian3DGaussian2024} & 0.774 & 0.945 & 0.868 & 0.629 & 0.790 & 0.781 & 0.916 \\
        & GSPRO \cite{chengGaussianPro3DGaussian2024} & 0.747 & 0.937 & 0.825 & 0.613 & 0.754 & 0.764 & 0.919 \\
        & OURS  & \textbf{0.793} & \textbf{0.950} & \textbf{0.891} & \textbf{0.644} & \textbf{0.830} & \textbf{0.791} & \textbf{0.923} \\ 
    \midrule
    \multirow{5}{*}{PSNR $\uparrow$}
        & 3D GS \cite{kerbl3DGaussianSplatting2023a}  & 24.99 & 31.19 & 27.11 & 20.16 & 26.97 & 23.31 & 25.49 \\ 
        & ABSGS \cite{yeAbsGSRecoveringFine2024} & 25.15 & 32.09 & 27.41 & 21.88 & 26.55 & 23.28 & 25.87 \\
        & RSGS \cite{yaoRSGaussian3DGaussian2024} & 25.48 & \textbf{32.25} & 27.56 & 22.22 & 27.04 & 23.06 & 25.77 \\
        & GSPRO \cite{chengGaussianPro3DGaussian2024} & 24.61 & 32.01 & 25.82 & 21.87 & 26.22 & 22.27 & 25.57 \\
        & OURS \cite{} & \textbf{25.88} & 32.24 & \textbf{28.72} & \textbf{23.81} & \textbf{29.77} & \textbf{25.26} & \textbf{28.99} \\
    \midrule
    \multirow{5}{*}{LPIPS $\downarrow$}
        & 3D GS \cite{kerbl3DGaussianSplatting2023a}  & 0.266  & 0.205  & 0.175  & 0.414  & 0.224  & 0.291  & 0.152  \\ 
        & ABSGS \cite{yeAbsGSRecoveringFine2024} & 0.237 & \textbf{0.181} & 0.123 & 0.348 & 0.246 & 0.250 & 0.140 \\
        & RSGS \cite{yaoRSGaussian3DGaussian2024} & 0.210 & \textbf{0.181} & 0.117 & 0.333 & 0.228 & 0.255 & 0.148 \\ 
        & GSPRO \cite{chengGaussianPro3DGaussian2024} & 0.218 & 0.221 & 0.143 & 0.345 & 0.244 & 0.265 & 0.137 \\ 
        & OURS  & \textbf{0.199} & 0.187 & \textbf{0.106} & \textbf{0.317} & \textbf{0.203} & \textbf{0.230} & \textbf{0.131} \\ 
    \bottomrule
    \end{tabular}% 
    }
\end{table*}

\subsection{Preliminary Knowledge for 3DGS}
\label{sec:preliminary}

3D GS initializes the point cloud as a set of 3D Gaussian ellipsoids using spherical harmonics functions related to the covariance matrix \(\sigma\), mean \(\mu\), opacity \(\alpha\), and color \(c\). Each Gaussian ellipsoid is defined by eq.\ref{eq:gaussian} , representing its shape and central position.

\begin{eqnarray}
	G(x)=e^{-\frac{1}{2}(x-\mu)^\top\Sigma^{-1}(x-\mu)}
	\label{eq:gaussian}
\end{eqnarray}

Its rasterization is obtained by \(\alpha\)-rendering:

\begin{eqnarray}
	C = \sum_i T_i \sigma_i c_i
	\label{eq:rendering}
\end{eqnarray}

where \(T_i\) is the opacity of the Gaussian ellipsoid, \(c_i\) is the color of the Gaussian ellipsoid.

The adaptive density control in 3D GS is key to ensuring an accurate representation of the scene, ensuring the destruction or cloning of geometry when positions are incorrect. 

\subsection{Spectral Residual Saliency Map Decoupling}
\label{sec:spectral}

Our low-frequency and high-frequency decoupling method is implemented through spectral residual saliency maps, and the pipeline is shown in Figure \ref{fig:workflow}.a. Through spectral analysis, our approach extracts residual high-frequency information from the frequency domain of the image and effectively decouples it from the low-frequency components.

Specifically, we first perform spectral analysis on the image using Fourier transform. To enhance the high-frequency components in the spectrum, we apply a logarithmic transformation to the spectral map, followed by a filtering technique to isolate the high-frequency part. The filter's standard deviation is learned through a Multi-Layer Perceptron (MLP), which is trained alongside the original scene geometry. Finally, by computing the spectral residual of the high-frequency region and masking the low-frequency region, we effectively decouple the scene's geometric structure from its texture details.

\begin{equation}
	H(x,y;\mathbf{S})=\frac{1}{2\pi\left(\mathrm{MLP}(\mathbf{S})\right)^2}\exp\left(-\frac{x^2+y^2}{2\left(\mathrm{MLP}(\mathbf{S})\right)^2}\right)
\end{equation}

\subsection{Low-Frequency Geometry Constraint}
\label{sec:low-frequency}

\begin{figure}[!ht]
	\centering
	\includegraphics[width=0.95\linewidth,keepaspectratio]{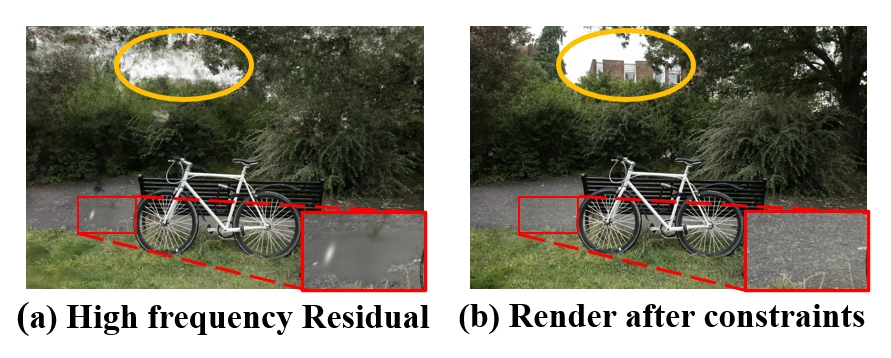}
	\caption{\textbf{Our Progressive Rendering Scheme.} (a) We render the high-frequency residual regions separately, while the surrounding geometric structure is not included (yellow ellipse). At this stage, some of the geometric structure is also problematic (red box). (b) The final result of the progressive rendering scheme. Not only have we accurately recovered the entire scene's geometric structure, but we have also effectively eliminated the artifacts present during separate rendering.}
	\label{fig:res}
  \end{figure}

The information in the low-frequency region primarily captures the geometric structure of the scene. In 3D Gaussian Splatting, misaligned Gaussian ellipsoids lead to the accumulation of color loss during rendering. To fit certain views, the properties of other Gaussian ellipsoids are learned incorrectly. At the same time, in an effort to fit as many views as possible, these ellipsoids become increasingly short, wide, and opaque, resulting in artifacts and floating objects in other views, as illustrated in Figure \ref{fig:arc}. Furthermore, the incorrect Gaussian ellipsoids may continue to split, ultimately affecting the scene's memory usage. To address this issue, we decouple the low-frequency region and, before restoring high-frequency texture details, first apply constraint-based corrections to the scene geometry in the low-frequency region.

\begin{figure*}[!t]
    \centering
    \includegraphics[width=0.9\textwidth, keepaspectratio]{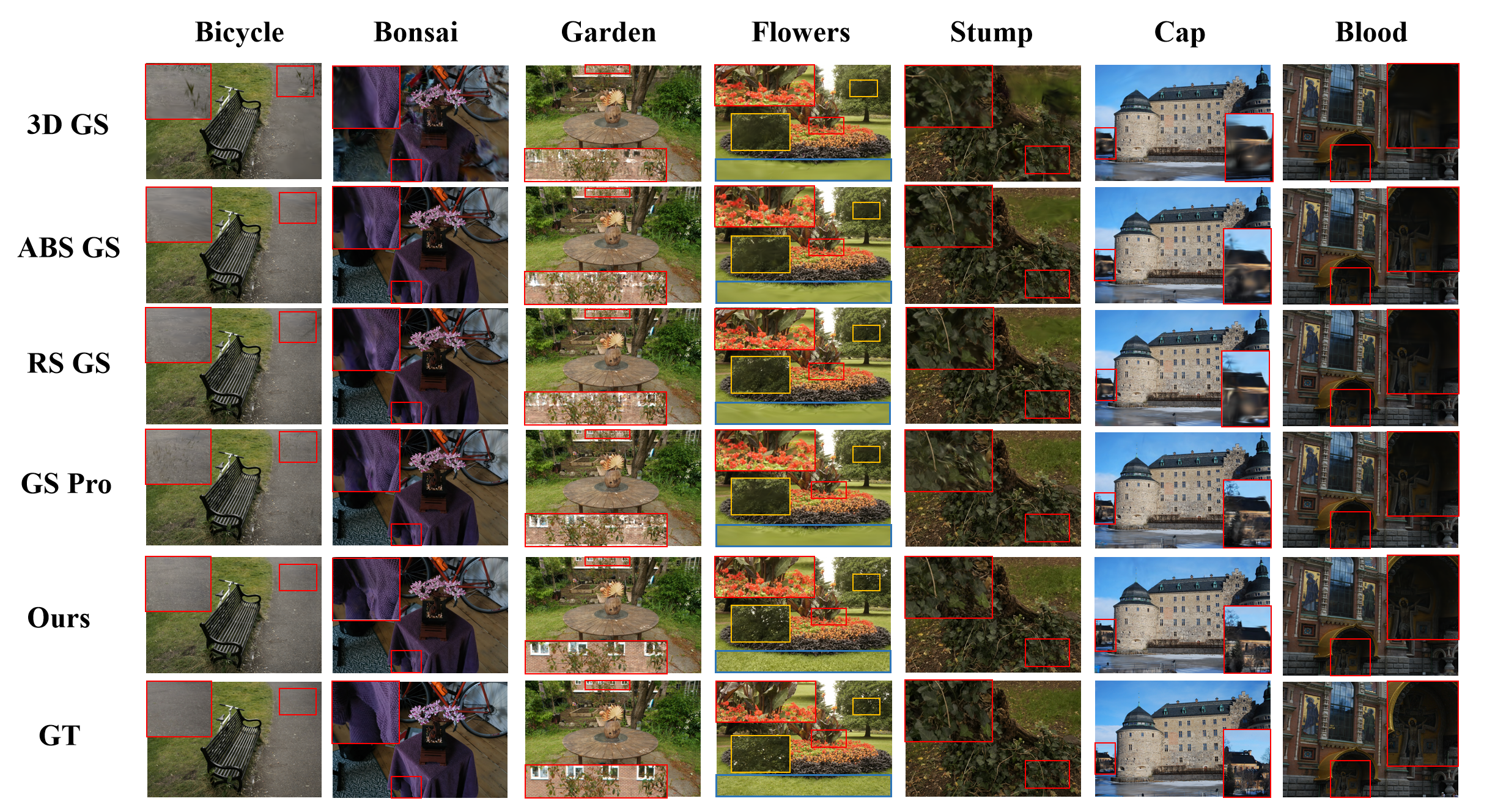}
    \caption{Visual comparisons with competing methods, showing that our approach better removes artifacts}
    \label{fig:output}
\end{figure*}

\noindent {\bf Depth map of 3D GS.} We modify the color \( c \) in the \( \alpha \)-rendering to depth \( d \), obtain the depth map through eq \ref{eq:rendering}.

\begin{equation}
	d^\text{alpha}=\sum_{i=1}^NT_i\alpha_id_i
	\label{eq:renderd}
\end{equation}

\noindent {\bf Normal map of 3D GS.} We adopt the method from Gaussian Pro \citep{chengGaussianPro3DGaussian2024}, where the direction of the shortest axis of the covariance matrix corresponds to the Gaussian normal direction, thus obtaining the normal map.

To correct the geometric structure, we combine Huber loss with an edge-aware depth loss, and introduce a depth-aware loss that is constrained using depth and normal maps from ground truth images across multiple views . Additionally, we apply a depth smoothing loss to prevent floating points and artifacts caused by errors in Gaussian ellipsoid fitting. As shown in eq \ref{eq:depth}.

\begin{equation}
	L_{\mathrm{dep}}=\sum_i\mathcal{L}_{\mathrm{Hu}}(d_i,\hat{d}_i)\cdot\mathrm{Edge}(i)+\lambda_3\sum_{\langle i,j\rangle}\left|\nabla d_i-\nabla d_j\right|^2
	\label{eq:depth}
\end{equation}

% Where the Huber loss behaves as L2 loss for small errors and as L1 loss for large errors, thereby increasing robustness to outliers. \( Edge \)  generated by an edge detection algorithm,ensures more accurate depth recovery in edge regions. By \( \sum_{\langle i,j\rangle}\left|\nabla d_i-\nabla d_j\right|^2 \)constraining the depth gradient differences between neighboring pixels, it reduces unnatural transitions or artifacts in the image caused by incorrect depth estimates.

\subsection{High-Frequency Texture Recovery}

In this process, we simultaneously render the constrained scene geometry and the original uncoupled scene geometry, and extract the gradient of the rendered images to obtain the corresponding gradient feature maps. By statistically analyzing the correlation coefficients between these gradient maps and the total variation loss, we ensure that the constrained scene geometry aligns with the original geometry and is densely optimized along the gradient direction. This guarantees that the high-frequency region has enough Gaussian ellipsoids to accurately fit the scene, thereby optimizing the recovery of fine details.

Next, we input the densified rendered image into a pre-trained network to compute the texture perception loss, further improving the texture quality of the details.

\section{Experiments}

Similar to other 3D GS variants, we evaluated and compared our method on the mip-NeRF 360 dataset \citep{barronMipNeRF360Unbounded2022} with multiple scenes. Additionally, to further validate the performance of our method on large-scale scenes, we used datasets from Lund University \cite{fb8a32a8c1964e04bd9f74f710b9650d} to demonstrate the superiority of our approach in terms of high-frequency detail recovery and the geometric accuracy of the final generated set of Gaussian ellipsoids. Following standard practice, we used the same hyperparameter settings as 3D GS, stopping Gaussian densification after 15k iterations and terminating training at 30k iterations. We set \( \lambda_3 \) to 0.01. We report new viewpoint synthesis metrics, including PSNR, SSIM, and LPIPS. The comparison results between various methods and our PSRGS are shown in Table \ref{table:comparison1}. We compare our approach with 3D GS \cite{kerbl3DGaussianSplatting2023a} and its variants \cite{yaoRSGaussian3DGaussian2024,yeAbsGSRecoveringFine2024,chengGaussianPro3DGaussian2024} from related fields. The results indicate that our method outperforms the others in most scenarios, and it consistently surpasses 3D GS. Additionally, we present our visualization results in the figures.

It is important to note that, in small scenes, the performance of our method is comparable to that of other methods. However, as mentioned, in large scenes, our method outperforms the others in terms of high-frequency details and the geometric accuracy of the generated Gaussian ellipsoid set. At the same time, our method is better at constructing sparse Gaussian scenes, avoiding the generation of artifacts. Both the quantitative metrics and the visualization results demonstrate the effectiveness of our approach.

\section{Conclusion}

In this paper, we have presented PSRGS, a progressive optimization framework designed to address the challenges of 3D Gaussian Splatting (3D GS) in large-scale remote sensing scenes. Our approach decouples the scene into low-frequency geometric structures and high-frequency texture details using spectral residual saliency maps, enabling separate and targeted optimization for each region. Both qualitative and quantitative evaluations confirm that PSRGS provides significant improvements in rendering quality, especially in high-frequency regions. In conclusion, PSRGS offers a promising solution for novel view synthesis, especially in large-scale remote sensing and other complex 3D scene applications.

\small
\bibliographystyle{IEEEtranN}
\bibliography{references}

\end{document}